\documentclass[10pt,twocolumn,letterpaper]{article}
\usepackage{iccv}
\usepackage{times}
\usepackage{epsfig}
\usepackage{graphicx}
\usepackage{amsmath}
\usepackage{amssymb}
\usepackage{xcolor}
\usepackage{soul} 
\usepackage{changes}
\usepackage{colortbl}
\usepackage{gensymb}
\usepackage{siunitx}
\usepackage{enumitem}

\usepackage[pagebackref=true,breaklinks=true,letterpaper=true,colorlinks,bookmarks=false]{hyperref}

\iccvfinalcopy 

\ificcvfinal\fi
\begin{document}

\title{Self-Supervised Spatiotemporal Feature Learning via Video Rotation Prediction
}

\author{
  Longlong Jing${^1}$ \quad Xiaodong Yang${^2}$ \quad Jinggen Liu${^3}$ \quad Yingli Tian${^1}$\thanks{Corresponding author.}\\ 
  $^1$The City University of New York ~$^2$NVIDIA Research ~$^3$JD AI Reserach\\
}



\maketitle

\begin{abstract}
The success of deep neural networks generally requires a vast amount of training data to be labeled, which is expensive and unfeasible in scale, especially for video collections. To alleviate this problem, in this paper, we propose 3DRotNet: a fully self-supervised approach to learn spatiotemporal features from unlabeled videos. A set of rotations are applied to all videos, and a pretext task is defined as prediction of these rotations. When accomplishing this task, 3DRotNet is actually trained to understand the semantic concepts and motions in videos. In other words, it learns a spatiotemporal video representation, which can be transferred to improve video understanding tasks in small datasets. Our extensive experiments successfully demonstrate the effectiveness of the proposed framework on action recognition, leading to significant improvements over the state-of-the-art self-supervised methods. With the self-supervised pre-trained 3DRotNet from large datasets, the recognition accuracy is boosted up by $20.4$\% on UCF101 and $16.7$\% on HMDB51 respectively, compared to the models trained from scratch. 
\end{abstract}

\section{Introduction}

With more videos flourishing on the internet, recognizing human actions from videos \cite{P3D, RC3D, videocapsulenet, UnsupervisedMeta, Sultani_2018_CVPR, multilayer, prernn} has drawn increasing attention in computer vision community. Recently, thanks to the strong capability of simultaneously capturing both spatial and temporal representations, 3DCNNs have been widely and successfully explored in many video understanding tasks \cite{I3D, T3D, 3DResNet,C3D}.

\begin{figure}
\begin{center}
\includegraphics[width=0.45\textwidth]{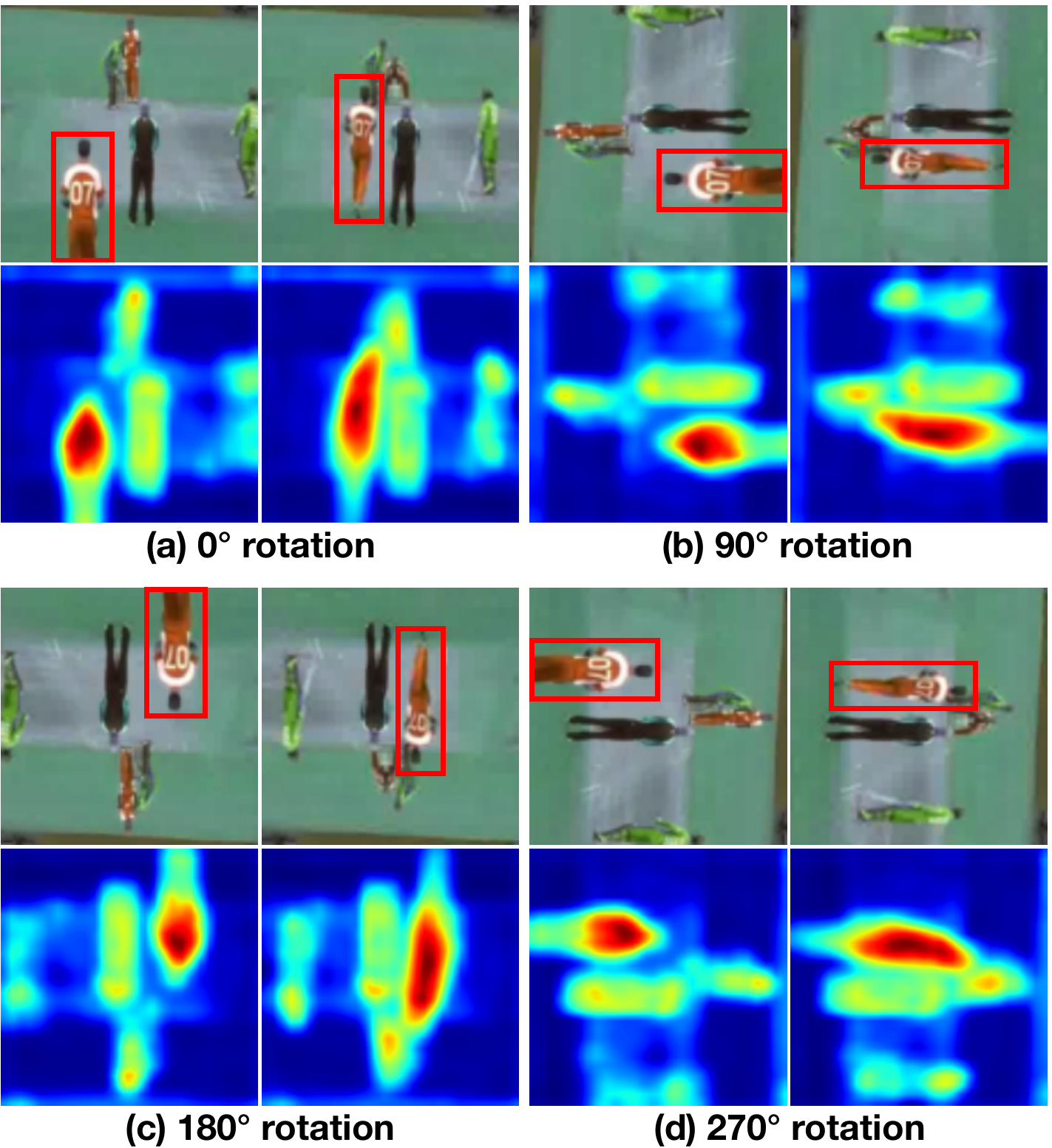}
\end{center}
\caption{Video frames and their corresponding attention maps generated by our proposed self-supervised 3DRotNet at each rotation. Note that both spatial (e.g. locations and shapes of different persons) and temporal features (e.g. motions and location changes of persons) are effectively captured. The hottest areas in attention maps indicate the person with the most significant motion (corresponding to the red bounding boxes in images). The attention map is computed by averaging the activations in each pixel which reflects the importance of that pixel.} 
\label{fig:motivation-attention}
\vspace*{-6mm}
\end{figure}

To achieve a good performance, 3DCNN-based supervised feature learning approaches require millions of video and label pairs for training. For instance, at the time C3D \cite{C3D} is proposed for action recognition on UCF101, a relatively small dataset, its performance is not comparable to that of hand-crafted features like dense trajectories \cite{IDT}. Until 3DCNNs are pre-trained on large-scale video datasets such as Sports-1M \cite{Sport-1M} and Kinetics \cite{Kinetics}, the performance has been largely improved. In fact, Kinetics consists of approximately $500,000$ videos of $600$ human actions. However, collecting such large-scale annotated video datasets is laborious and expensive in practice for new video understanding tasks. Therefore, here we attempt to learn spatiotemporal features directly from numerous unlabeled videos. 

To mitigate the aforementioned problem, in this paper, we propose 3DRotNet, a simple yet effective 3DCNN-based self-supervised spatiotemporal feature learning framework, which eliminates the requirement of human annotations. Our self-supervised learning defines an annotation-free pretext task to identify and provide supervisory signals solely from the visual information present in videos. This paradigm has been widely and successfully applied in image domain to learn image features for various image understanding tasks \cite{deepcluster, contextprediction, O3N, larsson2017colorproxy, shuffleandlearn,  jigsaw, watchingmove, inpainting, SynGAN}. As an example, in the pretext task of image inpainting, Pathak \textit{et al.} design a self-supervised 2DCNN to predict the missing regions in an image by learning the concept and the structure of the image \cite{inpainting}. Overall, the rationale behind the self-supervised approaches is that networks are enforced to learn high-level semantic features during accomplishing the pretext tasks. 

Following this learning strategy, our work is in particular designed to achieve a pretext task of recognizing the rotation transformation that is applied to videos, and as a result, it simultaneously learns a network capturing high-level semantic and motion features. As examples shown in Fig.~\ref{fig:motivation-attention}, in order to recognize how the video is rotated, a semantic sense to persons, objects as well as their locations and motions are needed. It is difficult to accomplish the pretext task without the knowledge of these semantic concepts. This is the underlying rationale behind our approach.  

Specifically, we first apply a set of rotations (e.g. $\ang{0}$, $\ang{90}$, $\ang{180}$ and $\ang{270}$) to videos as shown in Fig.~\ref{fig:motivation-attention}, and then define a pretext task as recognizing the set of rotations. To this end, the 3DRotNet is trained to recognize how many degrees each input video is rotated given raw frames (RGB) or difference of frames (DIF) of videos as inputs, where the latter can be treated as a light version of optical flow. During the training process, 3DRotNet attempts to learn a semantic video representation, which is able to capture spatial appearance cues (e.g. location and shape) as well as temporal information (e.g. motion and evolution). Fig.~\ref{fig:motivation-attention} illustrates the video frames and their corresponding attention maps generated by 3DRotNet at each rotation. It demonstrates the effectiveness of our approach to capture spatiotemporal video representations.

For quantitative evaluation, we pre-train 3DRotNet on Kinetics using the proposed self-supervised feature learning framework without annotations, and then transfer the learned features for action recognition tasks on UCF101 and HMDB51. The performance gap between our self-supervised feature learning and the supervised pre-trained models is getting small. In addition, our approach substantially outperforms other alternative self-supervised methods. A variety of ablation studies have been conducted to further analyze our models. 

Our work is inspired by some recent image-based self-supervised feature learning methods \cite{dosovitskiy2014discriminative, agrawal2015learning, rotation}, which also involve geometric transformations. The feature representations are yielded during the processes of learning to predict camera transformation using ego-motion~\cite{agrawal2015learning}, learning to be discriminative~\cite{dosovitskiy2014discriminative}, or learning to tell image rotation~\cite{rotation}. Instead, our work focuses on self-supervised video representation learning through simultaneous spatial and temporal feature modeling. 

In summary, our main contributions in this paper are as follows. First, we propose 3DRotNet, a simple yet effective fully self-supervised approach for spatiotemporal feature learning. By only using the video rotation transformation without requiring any annotations, 3DRotNet is capable of capturing both spatial and temporal information. Second, extensive experiments find our approach to produce significantly better results than the state-of-the-art self-supervised methods. Third, 3DRotNet learned in the unsupervised manner can be served as a pre-trained model to be transferred to other video understanding tasks when only small datasets are available. With 3DRotNet pre-trained on Kinetics dataset, the performance of action recognition is remarkably boosted up by $20.4\%$ on UCF101 and $16.7\%$ on HMDB51 compared to that from the models trained from scratch.       

\section{Related Work}

Recently, a number of self-supervised learning methods have been proposed to for representation learning from images and videos \cite{contextprediction, doersch2017multi, shuffleandlearn, mundhenk2018improvements, inpainting, SynGAN, SelfSurvey}. Based on the pretext tasks, these methods mainly fall into two categories: one is the texture based methods, which utilize texture information of images as supervision, such as the boundary of the objects \cite{unsupervisededges, SynGAN}, the context of images \cite{larsson2017colorproxy, inpainting}, and the similarity of two patches from an image \cite{contextprediction, contrasting, jigsaw, boosting, wang2017transitive}, and the other is the temporal based methods, which exploit temporal connection between frames such as the temporal order of frames \cite{O3N, OPN, shuffleandlearn} and cross-modal correspondence \cite{VTS, crossandlearn}.

\textbf{Self-Supervised Learning from Images.} The similarity between two patches from the same image is often used as a supervision signal for the self-supervised image feature learning \cite{contextprediction, contrasting, graphconstraint, jigsaw}. Noroozi and Favaro propose an approach to learn visual representations by solving Jiasaw puzzles with $9$ patches from same image \cite{jigsaw}. Doersch \textit{et al.} introduce to learn visual features by predicting the relative positions of two patches from the same image \cite{contextprediction}. Li \textit{et al.} propose to mine the positive and negative image pairs with graph constraints in the feature space and the mined pairs are used to train the network to learn visual features \cite{graphconstraint}. Caron \textit{et al.} present DeepCluster to iteratively train the network with categories that are generated by clustering \cite{deepcluster}. The context information of images such as structure \cite{inpainting}, color \cite{larsson2017colorproxy} and relations of objects is another type of supervision for self-supervised image feature learning. Gidaris \textit{et al.} propose to learn visual features by training a 2DCNN to recognize 2D image rotations which is proved to be helpful for image feature learning \cite{rotation}. Larsson \textit{et al.} employ image colorization as the pretext to learn semantic features of images \cite{larsson2017colorproxy}. Zhang \textit{et al.} present the split-brain autoencoder to predict a subset of image channels from other channels \cite{splitbrain}. Ren and Lee propose to learn image features from synthetic images generated by a game engine based on a generative adversarial network \cite{SynGAN}.

\textbf{Self-Supervised Learning from Videos.}
Although there are some prior work about self-supervised learning from videos, most of them still apply 2DCNNs to essentially learn image representations by using temporal information in videos as supervision. Pathak \textit{et al.} use a 2DCNN to segment moving objects that are unsupervised segmented from videos \cite{watchingmove}. Misra \textit{et al.} adopt a 2DCNN to verify whether a sequence of frames is in correct temporal order \cite{shuffleandlearn}. Wang and Gupta propose a Siamese-triplet network with a ranking loss to train a 2DCNN with the patches from a video sequence \cite{wang2015unsupervised}. Fernando \textit{et al.} propose to learn video representations by odd-one-out networks to identify the odd element from a set of related elements with a 2DCNN \cite{O3N}. Lee \textit{et al.} take shuffled video frames as input to a 2DCNN to sort the sequence \cite{OPN}.

3DCNN has been widely used to simultaneously model both spatial and temporal information in videos \cite{I3D, 3DResNet, 3D, P3D, C3D, VYOLO}, however, only a few attempts exploit it for self-supervised learning \cite{videogan,videocolorize, CubicPuzzles} and its performance is much lower than that of the supervised methods. Vondrick \textit{et al.} use a generative adversarial network for video generation and feature learning with a 3DCNN \cite{videogan}. They also propose to learn spatiotemporal features by colorizing videos with 3DCNN \cite{videocolorize}. Compared to the image based self-supervised learning, the spatiotemporal feature learning with 3DCNN is overall much less explored. 

\section{Method}

\begin{figure*}[!ht]
\begin{center}
\includegraphics[width=0.72\textwidth]{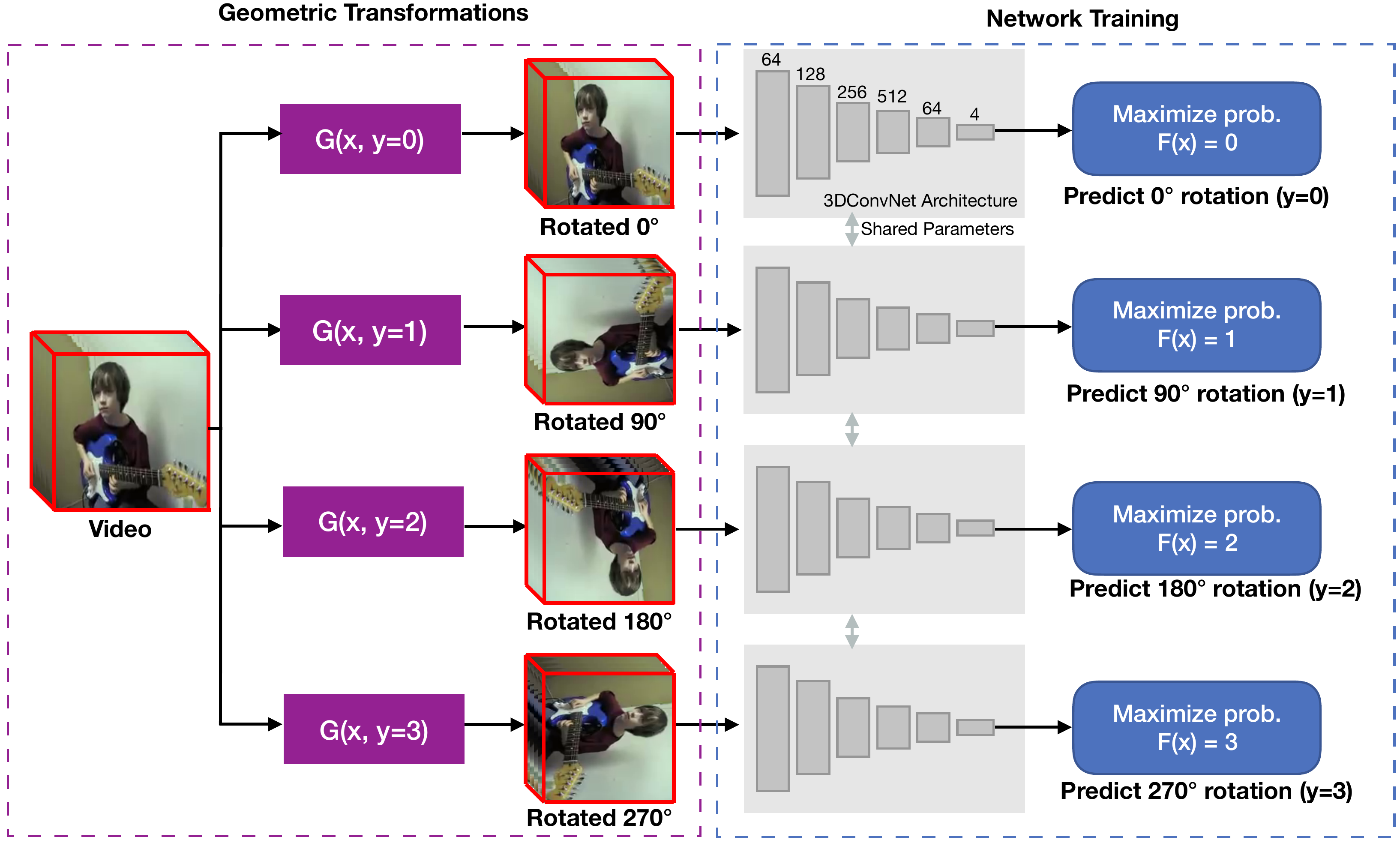}
\end{center}
\caption{The pipeline of the proposed self-supervised spatiotemporal representation learning. Each video is rotated with four different degrees ($\ang{0}, \ang{90}, \ang{180}, \ang{270}$), and 3DRotNet is trained to recognize the rotations that applied to input videos.}
\label{fig:FrameWork}
\vspace{-10pt}
\end{figure*}

\subsection{Model Parametrization}

In this paper, we adopt a 3DCNN $F(x|\theta)$ to learn the spatiotemporal features from predicting a set of pre-applied video rotation transformations $G(x|y) = Rot(x, y)$, while $x$ denotes the input video, $y$ indicates the parameter of the rotation transformation, the $Rot(x, y)$ is the rotation operation that rotates all the frames in a video with $y$ degrees and $G(x,y)$ is the result of transformation.

The video rotation prediction can be implemented in two ways: regression and classification. For the regression pretext task, the network predicts $y$ of the rotation transformation as a continuous variable. For the classification pretext task, a set of discrete rotations are pre-defined, then the network is trained to recognize the rotation category. 

Given a video $x_i$, the regression implementation is straightforward and can be formulated as: 

\begin{equation} 
\label{eq:loss_fun}
loss(x_i|\theta) = (F( G(x_i|y) | \theta) - y)^2,
\end{equation} 
The network $F(x|\theta)$ is trained to predict the parameter $y$ of the rotation transformation, while usually $\ell_1$ loss or $\ell_2$ loss is computed as the regression loss to optimize the network.



When formulate the problem as a classification task, a set of $K$ discrete rotations is defined, and the network $F(.)$ is optimized by minimizing the cross entropy loss between the predicted probability distribution over $K$ and the rotation $y$. The loss function is:
\begin{equation} 
\label{eq:loss_fun}
loss(x_i|\theta) = - \frac{1}{K}\sum_{y=1}^{K} log( F( G(x_i,y) | \theta)),
\end{equation}
In both scenarios, given  a  set  of $N$ training videos $D = \{x_i\}_{i=0}^{N}$, the overall training loss function is defined as:
\begin{equation} \label{eq:regression_loss}
loss(D) = \min_{\theta} \frac{1}{N}\sum_{i=1}^{N} loss(x_i|\theta).
\end{equation}

\subsection{Rotation Transformation Design} 

Different types of image transformations are designed as supervision information to train 2DCNNs for image representation learning including image colorization \cite{larsson2017colorproxy}, image rotation \cite{rotation}, and image denoise \cite{predictnoise}. In this paper, we propose to use video rotation as the supervision signal to learn video features. Specifically, 3DCNNs are trained to model both the spatial and temporal features which are representative for the semantic context of videos. Inspired by \cite{rotation, larsson2017colorproxy}, we formulate the problem as a classification task in which the network is to recognize $K$ types of discrete rotations that are applied to videos.

Choosing rotations as the geometric transformations for learning video features has the following advantages: (1) The problem is well-defined. Most of the videos in real-world environments are filmed in an upright way that the objects in the videos tend to be upright. (2) Compared to other pretext tasks, the rotation is easy to implement by the flip and transpose operations without adding much time complexity to the network. (3) Unlike other self-supervised learning methods need to take a lot of efforts to avoid the network to learn trivial solutions \cite{contextprediction}, the rotation operation leaves no artifacts in an image which can ensure the network learn meaningful semantic features through the process of accomplishing this pretext task. Following \cite{rotation}, we design four types of rotations at $\ang{0}$, $\ang{90}$, $\ang{180}$ and $\ang{270}$. Therefore, for each video $x$ with the type of rotation $y$, the output video after rotation transformation is $G = \{G(x|y)\}_{y=1}^{4}$, where $G(x|y) = Rot(x, (y-1)\times90)$.

\subsection{Proposed Framework}

Fig.~\ref{fig:FrameWork} illustrates the pipeline of the proposed 3DRotNet to learn spatiotemporal features by predicting the video rotation transformation. In our implementation, four kinds of rotations $G = \{Rot(X, (y-1)\times90)\}_{y=1}^{4}$ are applied to each video respectively. Then these four types of videos along with their rotation categories $y$ are used to train the 3DRotNet which predict the probability over all possible rotations for each video. The cross entropy loss is computed between the predicted probability distribution $F(X)$ and the rotation categories $y$ and is minimized to update the weights of the network. 

As for the network architecture, we follow the 3D ResNet18 since it has relatively fewer parameters and is capable to learn spatiotemporal features from large-scale datasets \cite{3DResNet}. There are five convolution blocks, while the first one consists of one convolution layer, one batch normalization layer, one ReLU layer, and followed by one max-pooling layer, and the rest four convolution blocks are 3D residual blocks with skip connection. The number of kernels in each convolution block is shown in Fig.~\ref{fig:FrameWork}. After the five convolution blocks, the global average pooling is applied to obtain a 512-dimensional feature vector. For rotation prediction, this 512-dimensional vector is followed by two fully connected layers with the dimensions of $64$ and $4$ to generate the prediction probability, while in the fine-tuning on action recognition task, this 512-dimensional vector is followed by only one fully connected layer of size equals to the number of action classes in the target action recognition dataset.

Unlike other self-supervised learning methods \cite{shuffleandlearn, jigsaw, watchingmove} that usually involve massive data preparation, our approach is straightforward to implement without complicated data preprocessing. Additionally, there is no extra effort needed to avoid trivial solutions since the rotation operations do not generate image artifacts. As a comparison in \cite{watchingmove}, masks of moving objects need to be generated in advance, and heavy data augmentation is applied to avoid the network to learn a trivial solution \cite{contextprediction}. 

\subsection{Evaluation Metrics}

To evaluate the quality of the learned features, previous self-supervised learning methods usually use the learned parameters as a start point and fine-tune on other high-level visual tasks such as action recognition. The performance of transfer learning on these tasks are compared to evaluate the generalization ability of the self-supervised learned features. If the self-supervised learning model can learn representative semantic features, then it can be served as a good start point and leads to better performance on these high-level visual tasks. In addition to the quantitative evaluation, previous methods also analyze network kernels and activation maps to provide qualitative visualization results \cite{deepcluster, contextprediction, shuffleandlearn, SynGAN}.

Following other self-supervised spatiotemporal feature learning methods such as \cite{CubicPuzzles}, our proposed approach is evaluated in the following ways. 

\begin{itemize}[noitemsep,topsep=0pt]
	\item Qualitatively analyze the kernels of the first convolution layer in 3DRotNet learned with the proposed approach and compare the kernels with that of the state-of-the-art supervised models.
	\item Analyze the feature activation maps generated by 3DRotNet and compare them with that of the state-of-the-art supervised models.
	\item Transfer the pre-trained 3DRotNet to action recognition task and compare the performance with the state-of-the-art self-supervised methods on two public benchmarks.
	\item Perform ablation studies to evaluate the impact of different configurations of the rotation transformation to the quality of the features learned by 3DRotNet.
\end{itemize}




\section{Experimental Results}

In this section, we conduct extensive experiments to evaluate the proposed approach and the quality of the learned spatiotemporal features for action recognition.

\subsection{Datasets}
Our self-supervised video representation learning network 3DRotNet is trained on two large-scale datasets Moment in Time \cite{MITS} and Kinetics \cite{Kinetics}. No action labels are used during the training. Following the standard evaluation protocol, the pre-trained 3DRotNet is then supervised fine-tuned for action recognition on two relatively small datasets: UCF101 \cite{UCF101} and HMDB51 \cite{HMDB51}, respectively. 

\textbf{Moment in Time (MT)} is a large-scale balanced and diverse dataset for video understanding \cite{MITS}. MT consists of around $1$ million videos covering $339$ action classes, and each video lasts around $3$ seconds. The average number of videos for each class is $1,757$ with a median of $2,775$. The training set of the dataset is used for the self-supervised learning without using video labels.

\textbf{Kinetics} is a large-scale and high-quality video dataset collected from YouTube \cite{Kinetics}. The dataset consists of around $500,000$ videos belonging to $600$ action classes with at least $600$ videos for each class. Each video is around 10 second. We download about $480,000$ videos and all of them are used to train our self-supervised model without knowing video labels.


\textbf{UCF101} is a widely used benchmark for action recognition. It consists of $13,320$ videos that cover $101$ human action classes. Due to the small size of the dataset, 3DCNNs often suffer from over-fitting on this dataset when trained from scratch. Pre-trained models (either supervised or self-supervised) from other large-scale datasets are needed to overcome over-fitting. 

\textbf{HMDB51} is another widely used benchmark for action recognition. It consists of $6,770$ videos in $51$ actions. Similar as UCF101, pre-trained models are also needed to alleviate over-fitting.

\subsection{Implementation Details}

\textbf{Self-Supervised Learning.} The videos in Kinetics and MT are evenly downsampled into $160$ and $90$ frames respectively and then are resized to a spatial resolution at $136 \times 136$. During training, $16$ (default value) consecutive frames are randomly selected from each video as a training clip, and a patch with $112 \times 112 $ pixels is randomly cropped from each frame to form a clip of size $3$ channels $\times$ $16$ frames $\times 112 \times 112$ pixels. Each video is horizontally flipped with $50$\% probability to augment the dataset. For each video, all the frames are rotated with four different degrees, and the four rotated videos are simultaneously fed into the network. The training is optimized by stochastic gradient descent (SGD) using $10,4000$ iterations and with a batch size of $32$. The initial learning rate is set to $0.1$ and is decayed by $0.1$ in every $2,4000$ iterations. 

\textbf{Transfer Learning.} To evaluate the learned features, we fine-tune the pre-trained model to perform action recognition on the two public datasets:  UCF101 \cite{UCF101} and HMDB51 \cite{HMDB51}. During training, $16$ consecutive frames are randomly selected from a video and resized to a spatial size of $136 \times 136$ pixels, then a $112 \times 112$ patch is cropped from each frame within the clip to form a tensor of size $3$ channels $\times$ $16$ frames $\times$ 112 $\times$ 112 pixels. The cross entropy loss is computed and optimized by SGD with $100$ epochs. The initial learning rate is set to $0.008$ and is multiplied by $0.1$ in every $4,000$ iterations. The top-1 classification accuracy on UCF101 and HMDB51 datasets are reported and compared.

\subsection{Can 3DRotNet Recognize Video Rotations?}

The hypothesis of our idea is that a network should be able to capture the semantic information in videos through recognizing video rotations, and the learned semantic information can be further transferred to other video understanding tasks such as action recognition. Therefore, we first test the performance of 3DRotNet to recognize the four rotations ($\ang{0}$, $\ang{90}$, $\ang{180}$ and $\ang{270}$). During training, the class labels of the videos in the two datasets (Kinetics \cite{Kinetics} and MT \cite{MITS}) are excluded and videos applied with the four rotations are used to train the 3DRotNet.

After trained on the two large-scale video datasets, the network is cross-domain tested on UCF101 and HMDB51. For testing, all videos in the two datasets are first applied with the four rotations, then the rotated videos are input to the 3DRotNet to predict the rotation angles. The average classification accuracy of the four rotations is shown in Table \ref{tab:rotation_acc}. The accuracy of video rotation recognition on UCF101 and HMDB51 are all higher than $89$\%, demonstrating that the proposed 3DRotNet is able to capture representative appearance cues in videos to recognize their rotations. However, it is still unclear whether the 3DRotNet can effectively capture the spatiotemporal information. 

\begin{table}[ht]
\begin{center}
\begin{tabular}{l|c|c}
\hline
Training Dataset   & UCF101 (\%)     & HMDB51 (\%)\\
\hline\hline
Kinetics           & $92.79$     & $93.66$\\
MT                 & $93.21$     & $89.88$\\
\hline
\end{tabular}
\end{center}
\caption{Accuracy of recognizing video rotations on  UCF101 and HMDB51 datasets. The 3DRotNet can accomplish this task with a accuracy of more than $89$\%.}
\label{tab:rotation_acc}
\end{table}

\subsection{Can 3DRotNet Learn Spatiotemporal Video Features?}

In order to verify whether the 3DRotNet learned from video rotations can capture both spatial and temporal features from videos such as moving objects, or whether the 3DRotNet solely rely a trivial solution such as using lines in videos to determine their rotations, we visualize the attention maps of the learned 3DRotNet models by averaging the activation maps of the first convolution layer, which can be used to reflect the importance of each pixel. 

\begin{figure}[t]
\begin{center}
\includegraphics[width=0.48\textwidth]{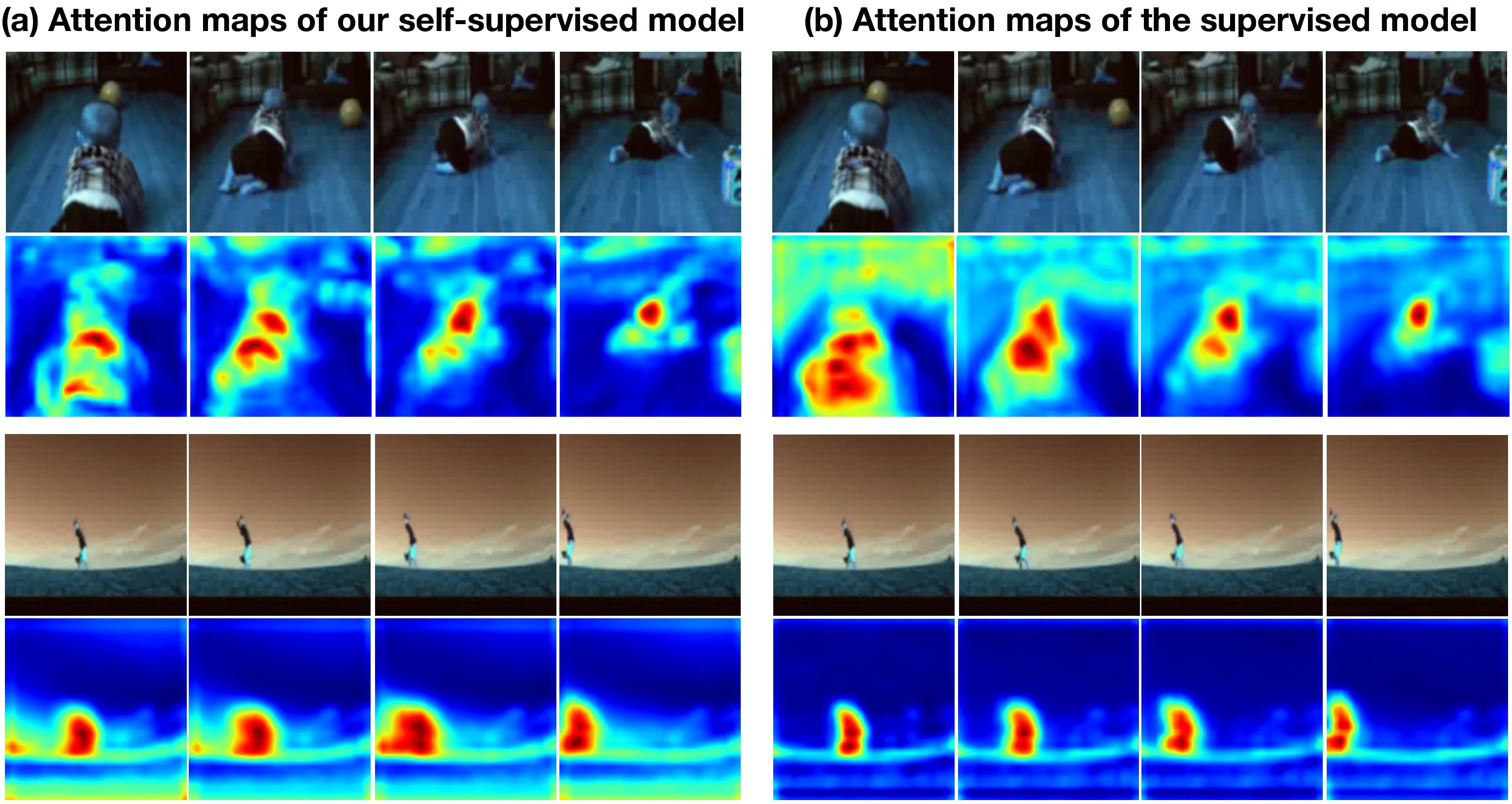}
\end{center}
\vspace{-5pt}
\caption{Sampled video frames and their corresponding attention maps generated by our proposed self-supervised 3DRotNet and by supervised model. The attention maps show that our model can capture both spatial and temporal information within videos. Moreover, the proposed self-supervised model can capture the main objects and their motions in a video as the supervised model.}
\label{fig:attentionmap}
\end{figure}

As shown in Fig.~\ref{fig:attentionmap}, the attention maps show that the 3DRotNet mainly focuses on the important foreground persons in videos and capture moving objects well. As shown in the images of the baby crawling video (right-bottom in Fig.~\ref{fig:attentionmap}), the 3DRotNet can capture the moving baby on the ground. This confirms that our 3DRotNet can capture the spatiotemporal information within videos.


\subsection{Transfer to Action Recognition Task}

In order to evaluate the generalization capability of the learned video features from our self-supervised models, we further conduct action recognition task on two different datasets (UCF101 and HMDB51) by using the learned video features (i.e., pre-trained 3DRotNet) as a start point and then finetuned on the action recognition datasets. The experimental results on the first split of UCF101 and HMDB51 datasets are shown in Table.~\ref{tab:transfer}.


\begin{table}[hb]
\begin{center}
\begin{tabular}{l|c|c}
\hline
Models & UCF101 (\%) & HMDB51 (\%) \\
\hline\hline
3DResNet (scratch) \cite{3DResNet} & 42.5   & 17.0\\
Ours (Kinetics)                & 62.9 \textbf{(+20.4)}  & 33.7 \textbf{(+16.7)}\\
Ours (MT)                      & 62.8 \textbf{(+19.2)}  & 29.6 \textbf{(+12.6)}\\
\hline
\end{tabular}
\end{center}
\caption{Results of transfer learning of the self-supervised model on action recognition task on UCF101 and HMDB51 datasets.}
\label{tab:transfer}
\end{table}


As shown in Table.~\ref{tab:transfer}, when the 3DResNet is trained from scratch on two action datasets it only achieves $42.5$\% on UCF101 and $17.0$\% on HMDB51 due to over-fitting. However, when fine-tune our self-supervised models on each dataset using the pre-trained models, the performance has a significant improvement of $20.4$\% (achieves $62.9$\%) on UCF101 and $16.7$\% ($33.7$\%) on HMDB51  which proves that the proposed self-supervised learning method is effective and indeed can provide a good start point for training a discriminative 3DResNets on the small datasets. Since MT dataset is very different to UCF101 dataset, the result on MT is more convincing. 

\begin{figure}
\begin{center}
\includegraphics[width=0.5\textwidth]{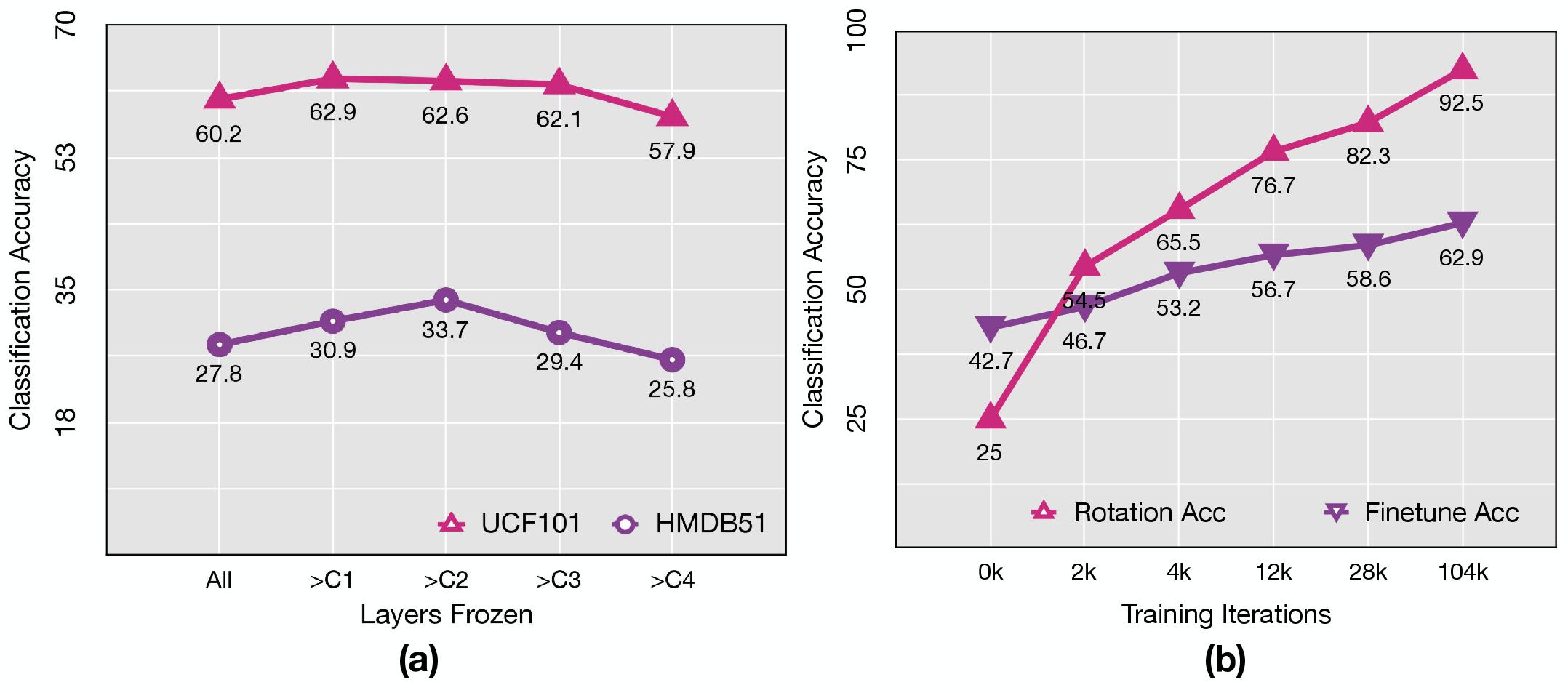}
\end{center}
\caption{(a) Finetune results on UCF101 and HMDB51 datasets. Cn means the n-th convolution block. $>$Cn means the blocks before the n-th convolution block are frozen during fine-tune. (b) Relation of the rotation recognition accuracy and the action recognition accuracy. The performance of action recognition increases along with the improvement of the accuracy of rotation recognition. }
\label{fig:rotation-acc}
\end{figure}

Following other self-supervised methods \cite{watchingmove}, the performance of CNNs layers frozen with different extent are compared and shown in Fig.~\ref{fig:rotation-acc} (a). The model pre-trained on Kinetics dataset is finetuned on HMDB51 and UCF101 datasets. For UCF101 dataset, the network has the best performance when the first convolution block is frozen, and has the worst performance when all the convolution blocks are frozen during training. For HMDB51 dataset, the network has the best performance when the first two convolution blocks are frozen, and has the worst performance when all the convolution blocks are frozen. This probably is because the lower layers learn the general low-level feature, while deeper layers learn the high-level task-specific features. When fine-tuned on the small dataset, the parameters of lower layers need to be preserved and deeper layers need to be further tuned for specific tasks. 

We also study the relationship between the accuracy of rotation recognition and the accuracy of action recognition on UCF101 dataset. The results are shown in Fig.~\ref{fig:rotation-acc} (b). The performance of action recognition increases along with the improvement of the accuracy of rotation recognition which validates that the proposed 3DRotNet can learn meaningful features for high-level video tasks through simple recognition of rotation geometric transformations.

\subsection{Ablation Study of Impact of Rotations}

We further conduct experiments to evaluate the impact of the combination of different rotation degrees to the accuracy of action recognition task under four situations: (a) Combining \ang{0} and \ang{90} rotations, (b) Combining \ang{0}, \ang{90}, and \ang{180} rotations, (c) Combining \ang{0}, \ang{90}, \ang{180}, and \ang{270} rotations, and (d) Combining \ang{90}, \ang{180}, and \ang{270} rotations. These networks are trained on Kinetics dataset and finetuned on UCF101 dataset. 

\begin{table}[ht]
\centering
\resizebox{0.45\textwidth}{!}{
\begin{tabular}{l c c c c}
 \hline
  Rotations & \multicolumn{4}{c}{Combination}\\ 
 \hline
  \hline
  \ang{0} rotation & \cellcolor{gray!25}$\surd$ & \cellcolor{gray!70} & \cellcolor{gray!25}$\surd$ & \cellcolor{gray!70} $\surd$ \\
  
   \ang{90} rotation & \cellcolor{gray!25}$\surd$ & \cellcolor{gray!70}$\surd$ &\cellcolor{gray!25} $\surd$&\cellcolor{gray!70}$\surd$ \\
   
   \ang{180} rotation & \cellcolor{gray!25}& \cellcolor{gray!70}$\surd$ & \cellcolor{gray!25}$\surd$ & \cellcolor{gray!70}$\surd$ \\
   
   \ang{270} rotation & \cellcolor{gray!25}& \cellcolor{gray!70} $\surd$ & \cellcolor{gray!25} & \cellcolor{gray!70}$\surd$ \\
   \hline
   Performance &\cellcolor{gray!25}50.94\% &\cellcolor{gray!70}58.79\% & \cellcolor{gray!25}59.24\% &\cellcolor{gray!70}62.90\%  \\
  \hline
\end{tabular}}
\vspace{2mm}
\caption{The comparison of the performance of networks to recognize different number of rotations on UCF101 dataset. The network that recognizes $4$ rotation degrees has the best performance among all the networks.} 
\label{tab:rotation_ablation}
\end{table}

Table~\ref{tab:rotation_ablation} illustrates the effects of the number of rotations to the transfer learning. The network trained for four rotations has the best performance on the transfer learning, and the network based  only two rotations has the worst performance. When only two kinds of rotations are available, the finetune performance on the UCF101 dataset is only $50.94$\% which is $11.96$\% lower than the performance of the network pre-trained with four rotations. This is probably because the network trained to recognize $4$ rotations received more supervision signal than the network trained to recognize $2$ rotations.

In addition to the 4-rotation recognition, we also train the 3DRotNet to recognize $8$ rotation degrees \{\ang{0}, \ang{45}, \ang{90}, \ang{135}, \ang{180}, \ang{225}, \ang{270}, and \ang{315}\} and to recognize $360$ rotation degrees. The 360-rotation network is optimized with regression loss. When fine-tuned on UCF101 dataset for action recognition, the 8-rotation network achieves only $57.0$\% performance which is $5.9$\% lower than that of the 4-rotation network, while the 360-rotation network achieves only $60.9$\% performance which is $2.0$\% lower than that of the 4-rotation network. The performance degradation probably comes from the context lost since only the center patch is cropped to avoid the empty image areas introduced by the rotation transformations.

\subsection{Ablation Study of Impact of Data Amount}

In this section, we evaluate the impact of the amount of training videos on the quality of features. We vary the amount of training data with Kinetics dataset for self-supervised learning and observe the action recognition performance of the transfer learning on action recognition task on UCF101 and HMDB51 datasets. The results are demonstrated in Table~\ref{tab:amount_ablation}.

\begin{table}[hb]
\begin{center}
\begin{tabular}{l|c|c}
\hline
Amount of Videos  & UCF101 (\%)  & HMDB51 (\%)\\
\hline\hline
$  0  $        &$42.5$     &$17.0$\\
$  100,000  $  &$56.4$ ($\textbf{+13.9}$)     &$27.3$ ($\textbf{+10.3}$)\\
$  200,000  $  &$58.3$ ($\textbf{+15.8}$)     &$28.1$ ($\textbf{+11.1}$)\\
$  400,000  $  &$60.7$ ($\textbf{+18.2}$)     &$28.6$ ($\textbf{+11.6}$)\\
$  All $       &$62.9$ ($\textbf{+20.4}$)     &$33.7$ ($\textbf{+16.7}$)\\
\hline
\end{tabular}
\end{center}
\caption{The relation of action recognition performance and the amount of training data used for self-supervised pre-training on Kinetics dataset. The performance keeps increasing as more data are used.}
\label{tab:amount_ablation}
\end{table}

As shown in Table~\ref{tab:amount_ablation}, the performance of the transfer learning increase as more training data is available which indicates that large-scale data is needed for self-supervised learning. The table also shows that the action recognition performance can be further improved by utilizing more training data for self-supervised pre-training.

\begin{table}[!h]
\begin{center}
\begin{tabular}{l|c|c}
\hline
Shots          & Scratch (\%)    & 3DRotNet (\%)\\
\hline\hline
1              & $8.33$           & $15.0$  ($\textbf{+6.7}$)\\
5              & $15.2$          & $31.5$  ($\textbf{+16.3}$)\\
10             & $19.9$          & $40.4$  ($\textbf{+20.5}$)\\
20             & $21.7$          & $47.1$  ($\textbf{+25.4}$)\\
Full           & $42.5$          & $62.9$  ($\textbf{+20.4}$)\\
\hline
\end{tabular}
\end{center}
\caption{Action recognition performance of few-shot learning on UCF101 dataset with and without using pre-trained models. When training dataset is extremely limited, the self-supervised pre-trained model can significantly improve the performance.}
\label{tab:few_shots}
\end{table}

In addition to the ablation study for the amount of training data for self-supervised pre-training in Table~\ref{tab:few_shots}, we conduct experiments to evaluate the performance of 3DRotNet when the training data for the target task is extremely small by few-shot learning on UCF101. Our self-supervised pre-trained model can significantly improve the performance when there are only a few samples are available. With 3DRotNet pre-trained model, even when only $20$ labeled videos are available, the performance of 3DRotNet for action recognition on UCF101 is comparable with the model trained with nearly $10,000$ labeled videos from scratch.

\subsection{Learning Long-Term Temporal Information}

In this section, we evaluate the impact of the length of input video clips as well as the quality of the features on RGB channel and the difference between frames respectively. We vary the length of input video clips for the self-supervised learning and observe the performance of the transfer learning on the action recognition task on the UCF101 dataset. The results are demonstrated in Table~\ref{tab:long_temporal}.

\begin{table}[!h]
\begin{center}
\begin{tabular}{l|c|c}
\hline
Lenght of Clip & UCF101 (\%)    & HMDB51 (\%)\\
\hline\hline
16-RGB         & $62.9$     & $33.7$\\
32-RGB         & $64.5$     & $34.3$\\
64-RGB         & $66.4$     & $37.4$\\
\hline
16-DIF         & $70.8$     & $40.0$\\
64-DIF         & $73.4$     & $42.0$\\
\hline
\end{tabular}
\end{center}
\caption{The comparison of the action recognition performance of networks with different length of input clips for both RGB and difference of frames (DIF). The networks with longer input clip achieve better performance for action recognition since long-term temporal information is provided by the video clip.}
\label{tab:long_temporal}
\end{table}

As shown in Table~\ref{tab:long_temporal}, the performance of the transfer learning increase as long input video clips are used. Simply by increasing the length of clips from 16 to 64, the performance increases 3.5\% on UCF101 dataset. The difference of frames (DIF) captures motion within video clips and invariant to appearance which probably leads to much higher performance than models trained with RGB clips. 

\subsection{Compare with Other Self-supervised Methods}

In this section, we compare our 3DRotNet with other self-supervised methods on action recognition task including the 2DCNN-based \cite{hadsell2006dimensionality, shuffleandlearn, mobahi2009deep, wang2015unsupervised, AOT} and the 3DCNN-based methods \cite{videogan, CubicPuzzles}. The 2DCNN-based methods mainly use the temporal information between frames as the supervision signal to train the 2DCNN. The features learned in most of these models are still focusing on the image features of every single frame \cite{wang2015unsupervised, shuffleandlearn}. However, the 3DCNN-based methods can simultaneously learn both spatial and temporal information in videos.

Table~\ref{tab:compare} shows the action recognition accuracy on UCF101 and HMDB51. The supervised models of 2DCNN and 3DCNN based methods have the state-of-the-art performance of over $80$\% on the UCF101 dataset \cite{3DResNet, I3D, P3D, TSN}. These models usually require pre-trained models from large-scale labeled datasets and involve fusion of different modalities such as the Optical Flow, RGB, and DIF.

Our 3DRotNet-RGB outperforms all the 2DCNN-based and 3DCNN-based self-supervised learning methods and achieves $0.2$\% and $3.7$\% higher on UCF101 and HMDB51 dataset respectively than the state-of-the-art self-supervised method \cite{CubicPuzzles}. The $Fusion$ indicates the geometric mean of RGB network and DIF network is computed to obtain the final score. The fusion boosts the performance by $2.3$\% and by $4.5$\% on the UCF101 and HMDB51 dataset. The gap between the fusion result and the supervised result \cite{3DResNet} is only $7.8$\% on UCF101 dataset. 

\begin{table}[t]
\small
\begin{center}
\begin{tabular}{c|c|c|c}
\hline
                &{Method}   &UCF101    &HMDB51\\
                &           &Acc (\%)  &Acc (\%)\\
\hline\hline
                            &C3D (Sport-1M)         &$82.3$   &---\\
{\scriptsize Supervised}    &3DResNet-18 (Kinetics)    &$84.4$   &$56.4$\\
                            &P3D (Kinetics)         &$84.4$   &---\\
\hline
                     &ObjectPatch  \cite{wang2015unsupervised} &$42.7$ &$15.6$\\
                     &TemporalCoherency  \cite{mobahi2009deep} &$45.4$ &$15.9$\\
                     &ShuffleLearn  \cite{shuffleandlearn}   &$50.9$ &$19.8$\\
{\scriptsize 2DCNN}  &{GeometryGuided}  \cite{Gan_2018_CVPR}   &$54.1$ &$22.6$\\
{\scriptsize Self-Supervised}    &{AOT} \cite{AOT}      &$55.3$  &---\\
                    &{OPN} \cite{OPN}      &$56.3$  &$22.1$\\
                    &{CrossLearn} \cite{crossandlearn}      &$58.7$  &$27.2$\\
                    &{O3N} \cite{O3N}         &$60.3$     &$32.5$\\
\hline

&3D AE \cite{CubicPuzzles}  &$48.7$  &---\\
&3D AE + future \cite{CubicPuzzles}  &$50.1$  &---\\
&3D inpainting \cite{CubicPuzzles}  &$50.9$  &---\\
{\scriptsize 3DCNN}  &VideoGAN \cite{videogan} &52.1 &---\\
{\scriptsize Self-Supervised}&3DCubicPuzzle \cite{CubicPuzzles}  &$65.8$  &$33.7$\\
            &Ours-RGB  &\textbf{66.0}   &\textbf{37.1}\\

&Ours-DIF       &$\textbf{74.3}$  &$\textbf{42.5}$\\
&Ours-Fusion    &$\textbf{76.6}$  &$\textbf{47.0}$\\
\hline
\end{tabular}
\vspace{3mm}
\caption{Comparison with other self-supervised methods on  action recognition task. Our proposed method outperforms all other self-supervised methods on both UCF101 and HMDB51 datasets. All the 3DCNN-based self-supervised methods use the same architecture, and all the accuracies are averaged over three splits.}
\label{tab:compare}
\end{center}
\vspace{-4mm}
\end{table}


\subsection{Kernel Comparison between Supervised and Self-supervised Models}

\begin{figure*}[!ht]
\begin{center}
\includegraphics[width=\textwidth]{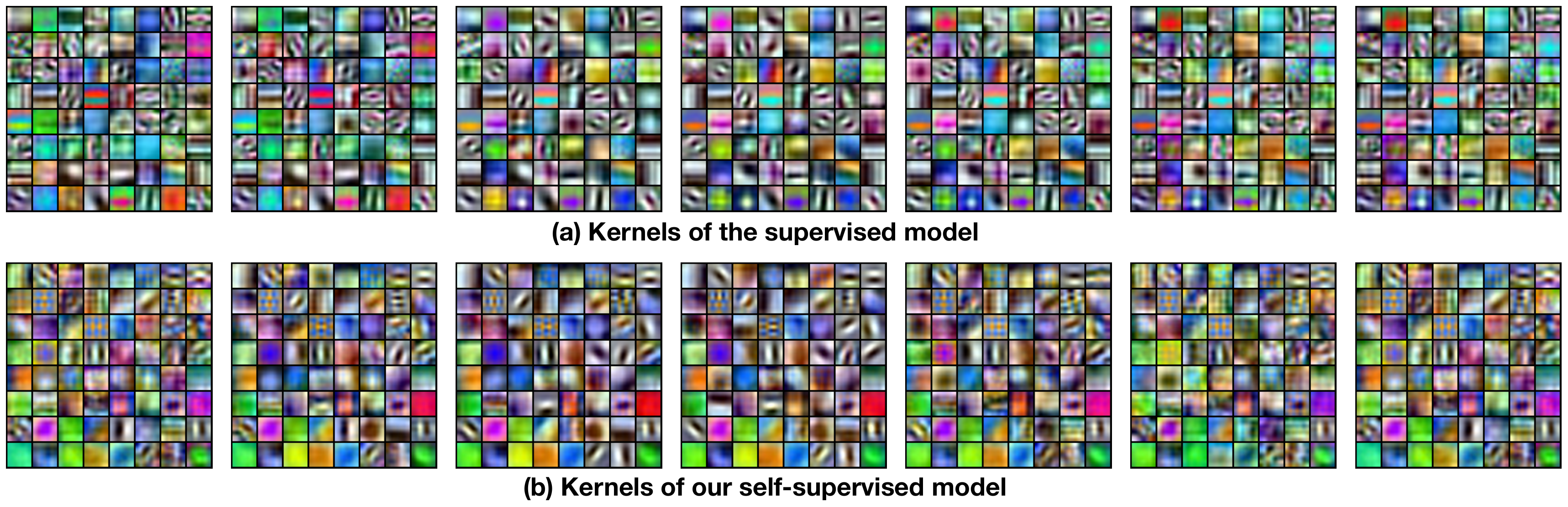}
\end{center}
\vspace{-10pt}
\caption{All the kernels of the first convolution block of our self-supervised 3DRotNet and fully supervised 3DResNet.}
\vspace{-10pt}
\label{fig:kernels}
\end{figure*}

Here, we visualize all the kernels of the first convolution layer of the proposed self-supervised 3DRotNet and the kernels of the fully supervised model in Fig. \ref{fig:kernels}. Both the proposed 3DRotNet and the supervised 3DConvNet are trained on the Kinetics dataset. The only difference is that the 3DRotNet is trained without the human-annotated category labels. As shown in Fig \ref{fig:kernels}, the self-supervised model learned the similar kernels as the supervised model.

\section{Conclusion}

We have proposed a straightforward and effective 3DCNN-based approach for self-supervised learning of spatiotemporal features from videos. The experiment results demonstrate that video rotation transformations are able to provide essential information for networks to learn both spatial and temporal features for videos. The effectiveness of the learned video features has been evaluated on action recognition task, and the proposed framework has achieved the state-of-the-art performance on two benchmarks among all existing self-supervised methods.

\vspace{3mm}

\noindent\textbf{Acknowledgement.}
This material is based upon the work supported by National Science Foundation (NSF) under award number IIS-1400802.


{\small
\bibliographystyle{ieee}
\bibliography{3DRotNet}
}

\begin{figure*}[!ht]
\begin{center}
\includegraphics[width=0.99\textwidth]{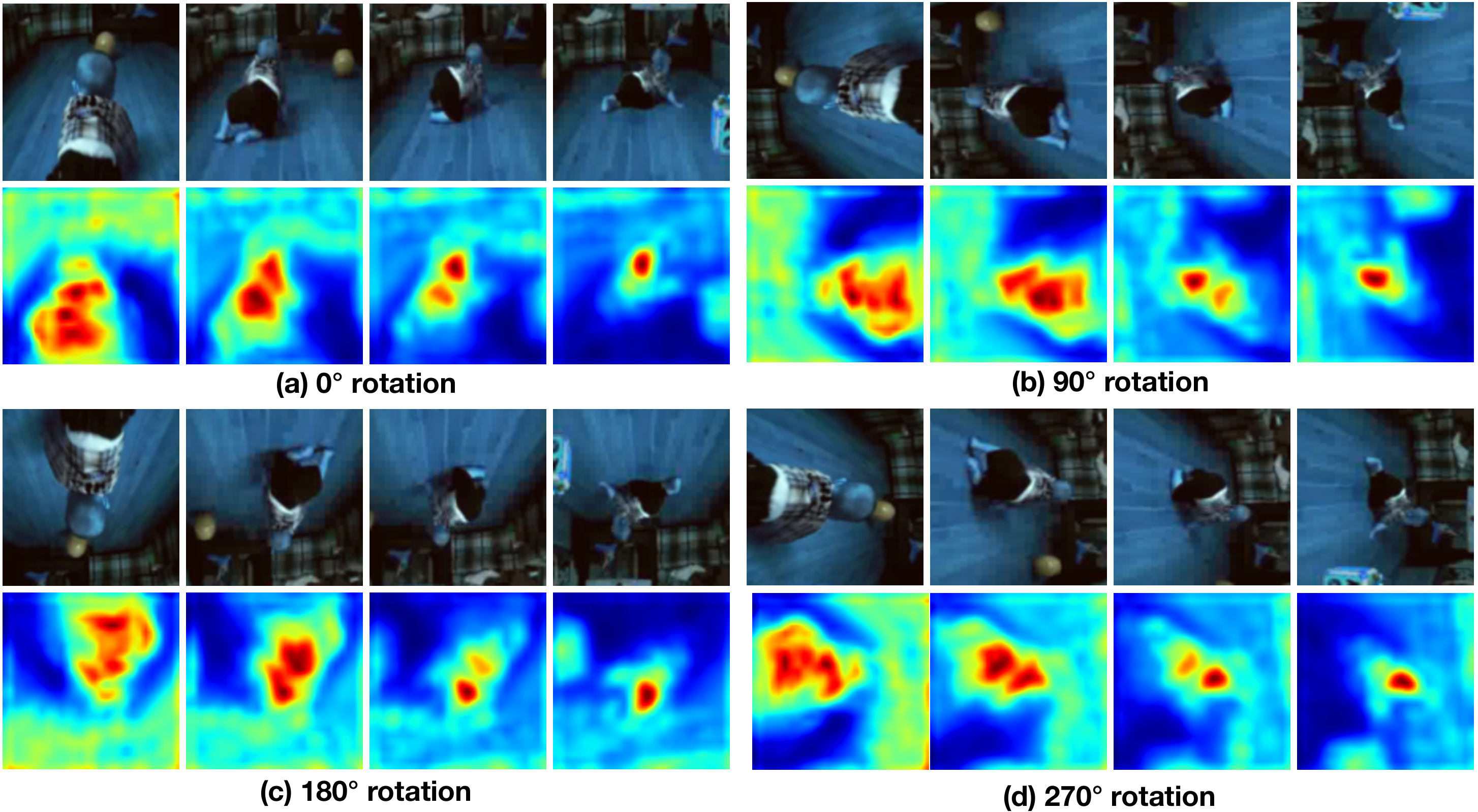}
\end{center}
\caption{Sampled video frames and their corresponding attention maps generated by our proposed self-supervised 3DRotNet model at each rotation angle. The network focuses on the moving baby at all rotations.}
\label{fig:311}
\end{figure*}

\vspace{60pt}

\begin{figure*}[!ht]
\begin{center}
\includegraphics[width=0.99\textwidth]{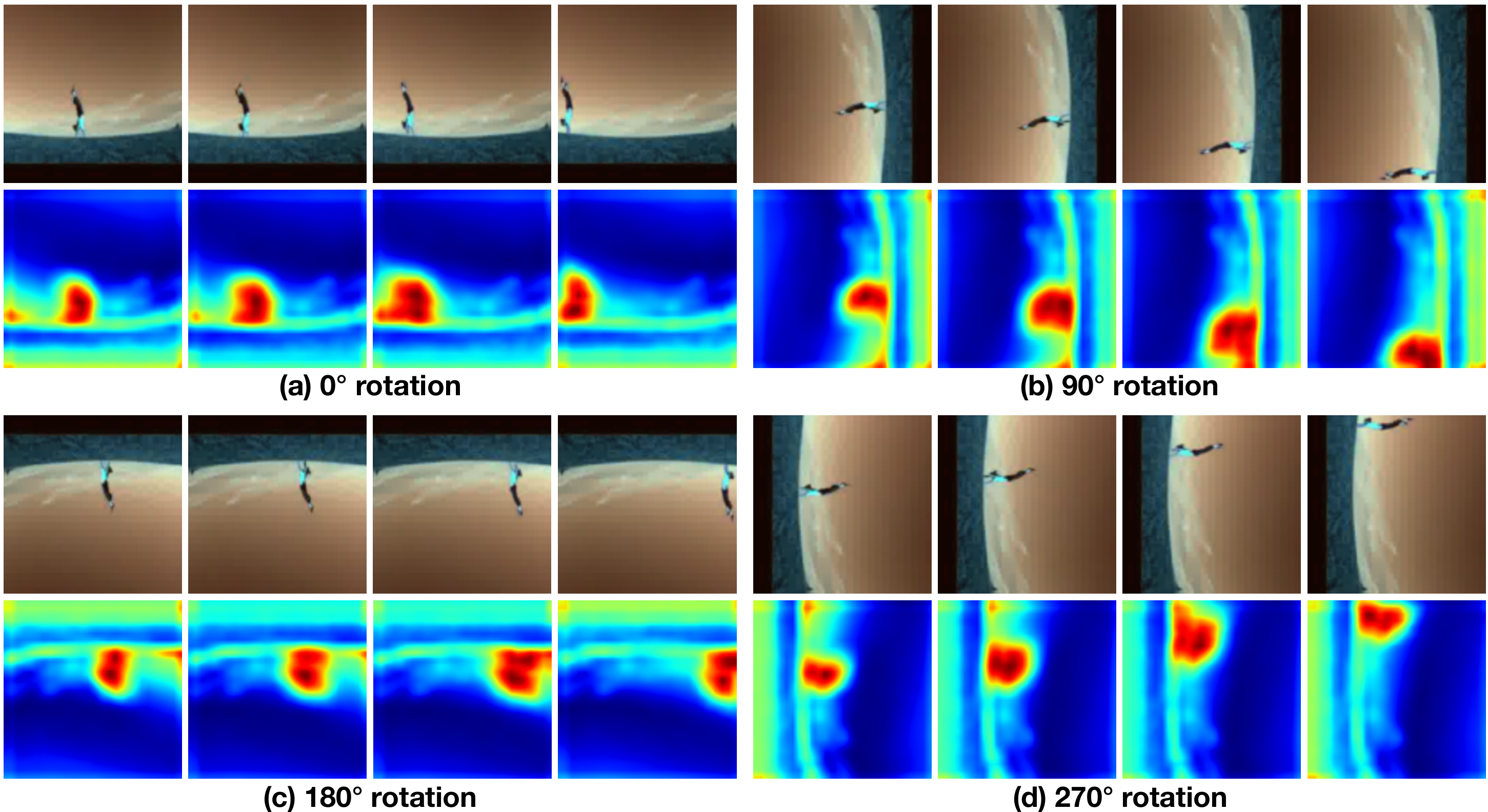}
\end{center}
\caption{Sampled video frames and their corresponding attention maps generated by our proposed self-supervised 3DRotNet model at each rotation angle. The network focuses on the moving person in this video.}
\label{fig:3607}
\end{figure*}

\begin{figure*}[!ht]
\begin{center}
\includegraphics[width=0.99\textwidth]{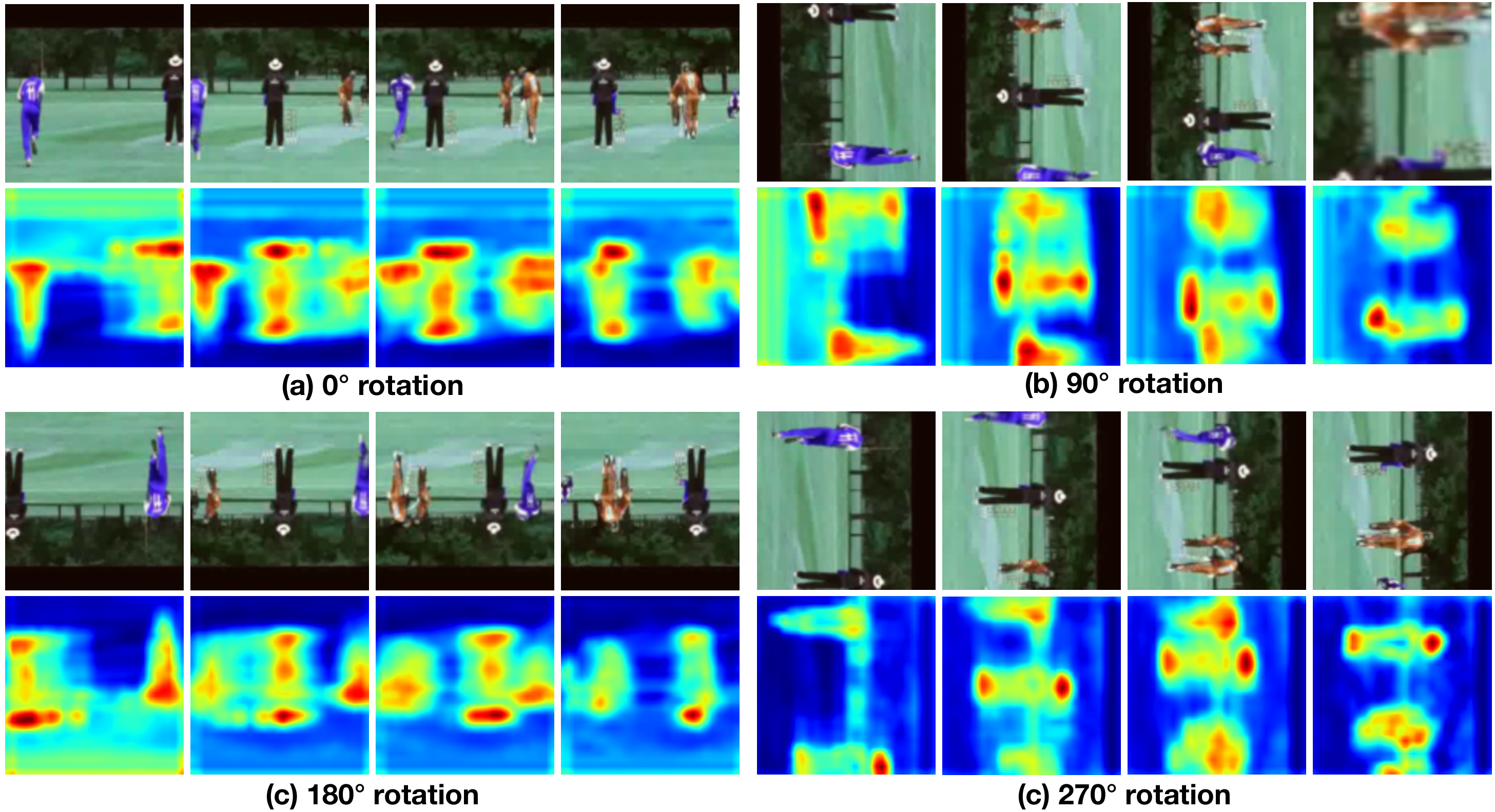}
\end{center}
\caption{Sampled video frames and their corresponding attention maps generated by our proposed self-supervised 3DRotNet model at each rotation angle. The network can capture the multiple persons at the same time among all the frames.}
\label{fig:2118}
\end{figure*}

\vspace{40pt}

\begin{figure*}[!ht]
\begin{center}
\includegraphics[width=\textwidth]{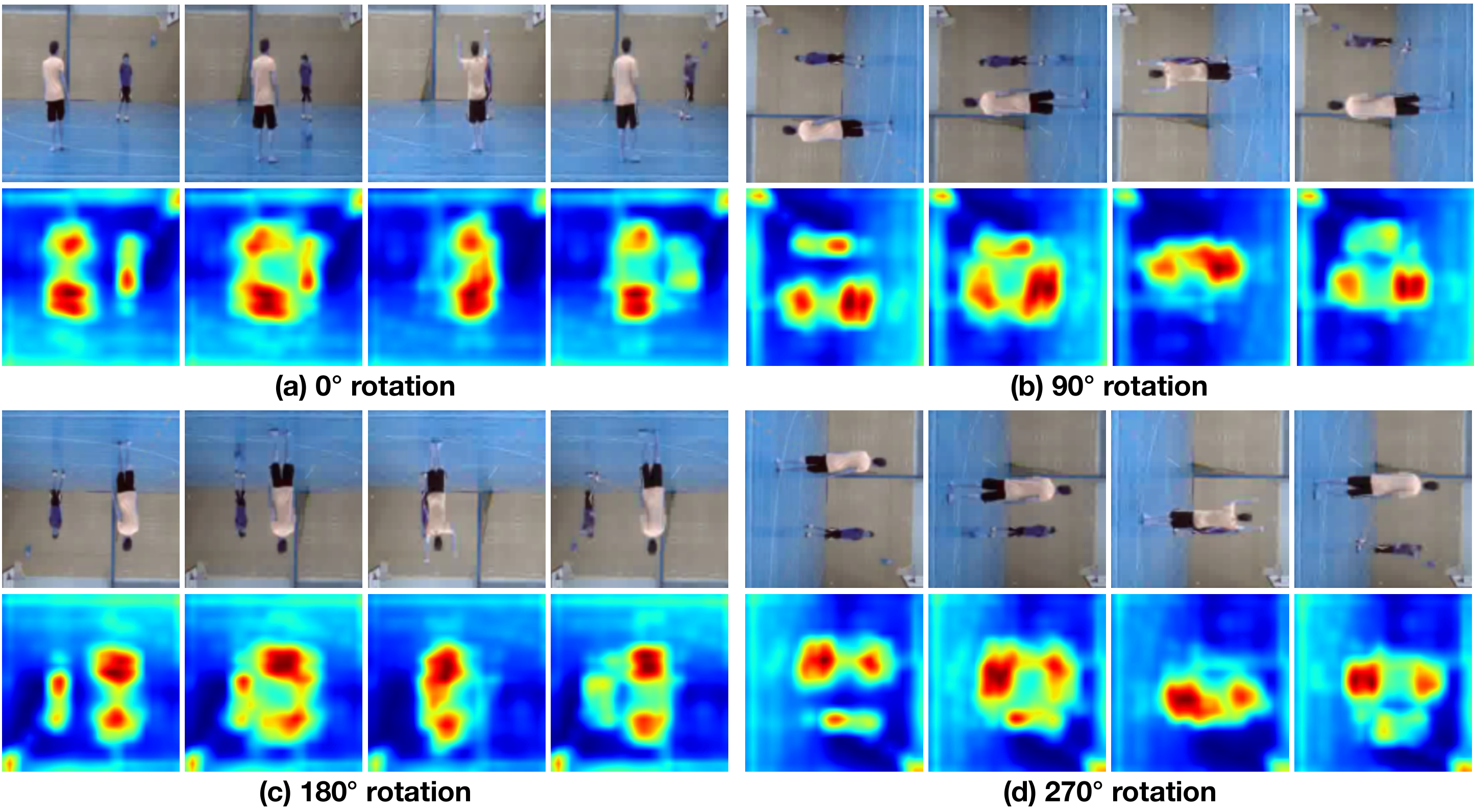}
\end{center}
\caption{Sampled video frames and their corresponding attention maps generated by our proposed self-supervised 3DRotNet model at each rotation angle. The network can capture the two persons at the same time and focuses on the person with the most significant movement.}
\label{fig:718}
\end{figure*}

\end{document}